%% file: main.tex
\documentclass{article}

\PassOptionsToPackage{numbers,compress}{natbib}
\usepackage[dblblindworkshop,final]{neurips_2026}
\workshoptitle{Verify-Agents Workshop}
\makeatletter
\renewcommand{\@notice}{}
\makeatother

\usepackage[utf8]{inputenc}
\usepackage[T1]{fontenc}
\usepackage{amsmath,amssymb,mathtools}
\usepackage{booktabs}
\usepackage{array}
\usepackage{multirow}
\usepackage{graphicx}
\usepackage{hyperref}
\usepackage{url}
\usepackage{xcolor}
\usepackage{microtype}
\usepackage{enumitem}
\usepackage{tikz}
\usetikzlibrary{arrows.meta,positioning,calc}

\hypersetup{
  colorlinks=true, linkcolor=blue, citecolor=blue, urlcolor=blue,
  pdftitle={Invocation-Level Reliability of Tool-Using Agents},
  pdfauthor={Afiya Noorain, Subhranshu Mohanty, Amritesh Banerjee, Abhijit Dasgupta}
}

\title{Invocation-Level Reliability of Tool-Using Agents}
\author{
  Afiya Noorain\textsuperscript{1}\thanks{Equal contribution.}\quad
  Subhranshu Mohanty\textsuperscript{1}\footnotemark[1]\quad
  Amritesh Banerjee\textsuperscript{2}\quad
  Abhijit Dasgupta\textsuperscript{3}\thanks{Corresponding author.} \\
  {\normalfont\textsuperscript{1}SAI International School, Bhubaneswar, Odisha, India} \\
  {\normalfont\textsuperscript{2}University of Massachusetts Amherst, Amherst, Massachusetts, USA} \\
  {\normalfont\textsuperscript{3}SP Jain School of Global Management, Mumbai, India}
}
\date{}

\begin{document}
\raggedbottom
\maketitle

\begin{abstract}
Tool-using agents fail two ways: choosing the wrong tool, or forming wrong arguments, and an early failure of either kind can silently corrupt everything downstream. We measure a correct-invocation rate that separates the two, under both a clean teacher-forced context and the model's own free-running context, on five open-weight models over contamination-free multi-step tasks (depths 1--8). By depth 6, roughly 70\% of a model's own clean-context capability is lost to its own earlier mistakes ($L_6 = 0.686, 0.684$). Our central finding concerns the measurement itself. \textbf{Under exact-match scoring against a fixed gold trajectory, a propagation model's severity and recovery parameters are not merely hard to estimate---they are fixed by the scoring rule.} Severity is forced to its boundary (0 of 869 poisoned steps correct); recovery is structurally unobservable (0 of 580 poisoned steps returned on-track, against an expected 0.0058 by chance). Both follow from one mechanism: post-divergence, the gold value is generated by tool constants the model never sees, so it is information the model cannot derive. A fit run anyway returns 0.92 and 0.73 for a quantity that is exactly 1.000---confident numbers for a parameter the scoring rule already determined. We give the mechanism and a remedy, conditional-on-state scoring, applied retrospectively to cached completions at zero additional cost, which un-pins severity to interior estimates excluding zero ($+0.149$, $+0.316$).
\end{abstract}

\section{Introduction}

A tool-using agent can reason correctly about what must happen and still fail the call, either by choosing the wrong tool or filling its arguments wrongly. Single aggregate success scores do not say which occurred, and typically reflect clean single-shot conditions rather than the multi-step, state-carrying tasks agents actually run. A model whose per-call accuracy looks fine in isolation can still fail badly in a chain once an early mistake corrupts everything depending on it.

We measure reliability at the level of the individual invocation: correct when the call both selects the right tool and supplies correct arguments, scored under a clean teacher-forced context (isolating how reliability degrades as context accumulates) and under the model's own free-running context (where an earlier mistake poisons what follows). The gap between the two is a direct signature of propagation, separable from ordinary context-length decay.

The study began as an empirical comparison and produced a measurement result instead. An interpretable severity/recovery model could not be fit to the data for a specific and general reason: the scoring rule, not the data, determines both parameters. We report the empirical findings and that result together, since the second explains why the first must be read through a fit-free metric.

\paragraph{Contributions.}
\begin{enumerate}[leftmargin=*,itemsep=1pt,topsep=1pt,parsep=0pt]
  \item A disaggregated correct-invocation protocol---matched clean/free-running scoring, syntactic-vs-semantic classification, and a routing-task design in which \emph{selection} errors propagate.
  \item A negative result about a scoring regime, not a model family: under exact-match scoring, severity and recovery are determined by the scoring rule, not the data. We give the mechanism and scope (any execution-match or AST-match benchmark).
  \item A remedy---conditional-on-state scoring---implemented and applied retrospectively to cached completions at zero marginal cost.
  \item Two per-model diagnostics invisible to aggregate accuracy: a \emph{discrimination} statistic (one model is anti-correlated with the rule it is given) and \emph{error position}, which bounds how much propagation a fixed-depth measurement can even observe.
  \item A released pipeline with measured provider rate limits documented nowhere else.
\end{enumerate}

\section{Related Work}

Toolformer~\cite{schick2023toolformer} learned when and how to call an API; ReAct~\cite{yao2023react} established the reason--act--observe loop most agent frameworks still follow; Gorilla~\cite{patil2023gorilla} first measured API-call correctness at scale, counting hallucinated calls as a distinct error.

Most large benchmarks score task success rather than per-call correctness. ToolLLM~\cite{qin2024toolllm} and StableToolBench~\cite{guo2024stabletoolbench} evaluate over thousands of (simulated, cached) APIs; API-Bank~\cite{li2023apibank} scores by execution match; BFCL~\cite{patil2025bfcl} combines AST matching with executable checks; tau-bench~\cite{yao2024taubench} grades final environment state, where even strong models succeed on fewer than half of tasks and are inconsistent across runs. \textbf{All of these score against a fixed reference}, the property Section~\ref{sec:central-result} shows to be limiting.

A smaller group disaggregates: RoTBench~\cite{ye2024rotbench} separates selection, parameter identification, and content filling under increasing noise; MTU-Bench~\cite{wang2025mtubench} reports selection and parameter accuracy without an LLM judge; FuncBenchGen~\cite{maekawa2025funcbenchgen}, the closest prior work, generates contamination-free multi-step dependency-graph tasks and finds capable models carrying stale values forward as chains lengthen---though it reports task success, not a disaggregated invocation measure.

Failures sort into \textbf{selection errors} (wrong/invented/omitted tool) and \textbf{argument errors} (malformed, missing, wrong, or fabricated values). Relign~\cite{xu2025relign} formalises this split and adds deferral; Hammer~\cite{lin2024hammer} masks function/parameter names for generalisation; representation-level detection~\cite{healy2026internal} flags bad selection at inference time. \textbf{These target the rate an incorrect call is produced, not what happens to a chain once one has entered it}---the gap this work addresses.

\section{Method}

\begin{figure}[t]
\centering
\resizebox{0.98\textwidth}{!}{%
\input{methodology_tikz.tex}%
}
\caption{The two-arm scoring protocol (A) and the three-state propagation model it measures (B). Section~\ref{sec:central-result} shows that under exact-match scoring, panel B's severity and recovery parameters are forced to the values shown, independent of the true underlying values.}
\label{fig:methodology}
\end{figure}
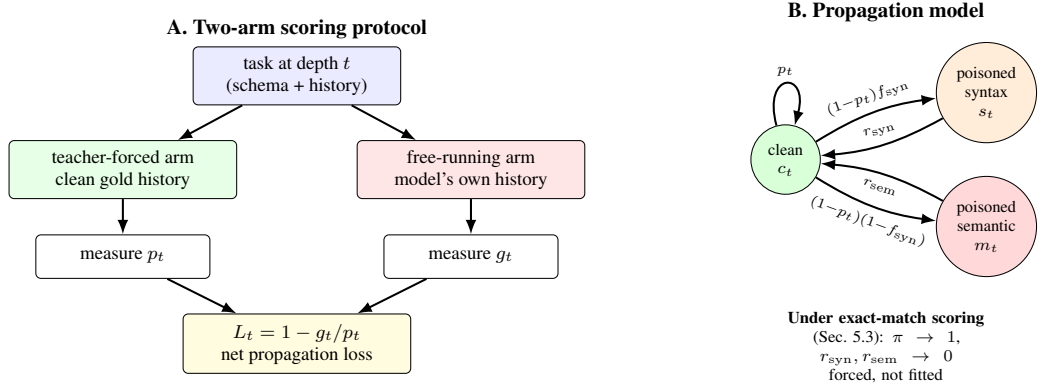

\subsection{Definitions}
\label{sec:definitions}

A task is a sequence of calls $c_1,\dots,c_n$ where the arguments of $c_t$ may depend on outputs of earlier calls; dependency depth is the length of the longest such chain.

\begin{itemize}[leftmargin=*,itemsep=1pt,topsep=1pt,parsep=0pt]
  \item \textbf{Local correctness $p_t$}: probability step $t$ is correct given a correct upstream history of length $t-1$ (teacher-forced). Measured at every depth, so it captures context-length degradation on its own.
  \item \textbf{Global correctness $g_t$}: probability step $t$ is correct inside the model's own free-running trajectory, where upstream inputs may already be poisoned.
  \item \textbf{Net propagation loss} $L_t \coloneqq 1 - g_t/p_t$.
\end{itemize}

\textbf{$L_t$ is defined purely empirically, from two measured rates, and nothing in this paper relies on any parametric relationship between $L_t$ and a model's severity.} Section~\ref{sec:propagation-model} introduces a three-state model under which $L_t$ \emph{would} equal $\pi \cdot x_t$ for a severity parameter $\pi$ and poisoned mass $x_t$---but Section~\ref{sec:central-result} shows that under exact-match scoring $\pi$ is forced to $1$ regardless of a model's true severity, which makes that identity vacuous as an estimation tool. We keep $L_t$'s definition and its parametric interpretation strictly separate for this reason: every empirical claim in Section~\ref{sec:results} uses only the definition above.

\subsection{A propagation model with recovery}
\label{sec:propagation-model}

We additionally model context as a three-state Markov process---clean, poisoned-by-syntax, poisoned-by-semantics (Figure~\ref{fig:methodology}B)---driven by measured $p_t$ and measured syntactic error share $f_{\mathrm{syn}}(t)$, with origin-specific recovery rates $r_{\mathrm{syn}}, r_{\mathrm{sem}}$ (syntactic failures are caught by the executor; semantic failures execute silently). A severity parameter $\pi$ scales correctness while poisoned: $\rho_t = (1-\pi)p_t$. Occupancies evolve as
\begin{align}
c_{t+1} &= c_t\,p_t + s_t\,r_{\mathrm{syn}} + m_t\,r_{\mathrm{sem}}, \quad
s_{t+1} = c_t\,(1-p_t)\,f_{\mathrm{syn}}(t) + s_t\,(1 - r_{\mathrm{syn}}), \\
m_{t+1} &= c_t\,(1-p_t)\,(1-f_{\mathrm{syn}}(t)) + m_t\,(1 - r_{\mathrm{sem}}),
\end{align}
with $g_t = p_t\,(1 - \pi\,x_t)$, poisoned mass $x_t = s_t + m_t$, so that \emph{under this model} $L_t = \pi\,x_t$. We fit $\pi, r_{\mathrm{syn}}, r_{\mathrm{sem}}$ hierarchically (PyMC, NUTS), reporting $\hat R$, ESS, divergences, and---critically---the posterior correlation between $\pi$ and each recovery rate, the diagnostic for separable identification used in Section~\ref{sec:central-result}.

\subsection{Task design}

Each task is a hidden dependency graph over deterministic executable functions, giving exact ground truth per call and direct control of depth ($\{1,2,4,6,8\}$). Tool outputs are $(a \cdot \mathrm{arg} + b) \bmod M$ with per-tool constants $a,b$ \textbf{never exposed in the schema}---the detail driving Section~\ref{sec:central-result}.

\textbf{Routing tasks (primary)}: the correct next tool is a parity rule applied to the incoming value, not a pre-announced order, so a wrong choice changes what is correct downstream. Selection is scored \emph{conditionally}---against the tool correct given the value the model actually holds---with divergence from gold recorded separately, so applying the rule correctly to a poisoned input is not misclassified as selection failure. \textbf{Linear tasks (control)}: order announced, argument copied verbatim. Degenerate on real models ($p_t=g_t=1.000$, both llama models); retained as a null design but \textbf{not executed} (Section~\ref{sec:limitations}).

\subsection{Protocol and suite}

We score selection conditionally and arguments by exact match (digit-only JSON strings coerced to integers, so provider formatting is not scored as a value error); classify errors as syntactic (parse/schema failure) or semantic (executes but wrong); measure $p_t$/$g_t$ as above; measure recovery from logs as the rate a poisoned context returns on-track. Both arms use the \textbf{same within-step retry budget}, so their ratio is not a retry artifact.

Five open-weight models, Groq free tier: \texttt{llama-3.1-8b-instant}, \texttt{allam-2-7b}, \texttt{qwen3.6-27b}, \texttt{gpt-oss-20b}, \texttt{gpt-oss-120b}, \texttt{llama-3.3-70b-versatile}. Greedy decoding; every response cached by (model, calling mode, exact prompt); fixed seeds. \textbf{Design deviations, all forced by free-tier hosting}: two scale points per family rather than 3--4 (no free host offers more); no proprietary reference (our Gemini key returns 403 on \texttt{generateContent}, confirmed against raw REST); no FC-tuned-vs-base contrast (no free-hosted xLAM/Hammer variant). All model IDs were verified live before running---the Qwen2.5/Llama-3.1-instruct generation used in our original design had since been retired from this tier.

Models run \textbf{nested prefixes} of one task suite, sized per model to its own measured token allowance. Task identity depends only on $(\mathrm{depth},k,\mathrm{seed})$, never on how many tasks were requested, so cross-model contrasts on the shared prefix remain exactly paired while each model's own estimate uses its full $n$.

\section{Data Collected}
\label{sec:data}

All figures come from a dataset \textbf{frozen at 14:09 on 2026-08-17}; the analysis is a pure function of the frozen files, with nothing extrapolated.

\begin{table}[htbp]
\centering
\caption{Per-model sample sizes at the freeze.}
\label{tab:samples}
\footnotesize
\begin{tabular}{@{}lrrrrrrl@{}}
\toprule
Model & tasks & $d_1$ & $d_2$ & $d_4$ & $d_6$ & $d_8$ & status \\
\midrule
\texttt{llama-3.1-8b-instant} & \textbf{273} & 60 & 60 & 65 & 65 & 23 & only model at depth 8 \\
\texttt{allam-2-7b} & \textbf{225} & 60 & 60 & 65 & 40 & --- & usable \\
\texttt{gpt-oss-120b} & \textbf{108} & 26 & 26 & 28 & 28 & --- & usable \\
\texttt{qwen3.6-27b} & \textbf{98} & 26 & 26 & 28 & 18 & --- & usable \\
\texttt{llama-3.3-70b-versatile} & \textbf{49} & 12 & 12 & 13 & 12 & --- & usable, thin \\
\texttt{gpt-oss-20b} & \textbf{0} & --- & --- & --- & --- & --- & \textbf{excluded} \\
\bottomrule
\end{tabular}
\end{table}

\texttt{gpt-oss-20b} completed zero tasks (pilot runs consumed its daily allowance) and is excluded from all claims. The binding constraint is an undocumented daily token allowance (TPD), non-uniform across models ($5\times$ spread), readable only from 429 response bodies.

\begin{table}[htbp]
\centering
\caption{Measured per-model rate limits, Groq free tier.}
\label{tab:tpd}
\footnotesize
\begin{tabular}{@{}lrrr@{}}
\toprule
Model & TPD (measured) & RPD & TPM \\
\midrule
\texttt{llama-3.1-8b-instant} & $>500{,}000$ (not reached) & 14{,}400 & 6{,}000 \\
\texttt{allam-2-7b} & \textbf{500{,}000} & 7{,}000 & 6{,}000 \\
\texttt{gpt-oss-20b}/\texttt{-120b}, \texttt{qwen3.6-27b} & 200{,}000 & 1{,}000 & 8{,}000 \\
\texttt{llama-3.3-70b-versatile} & \textbf{100{,}000} & 1{,}000 & 12{,}000 \\
\bottomrule
\end{tabular}
\end{table}

\section{Results}
\label{sec:results}

\subsection{Propagation is large, monotone in depth, and separated from context decay}
\label{sec:propagation-large}

\begin{figure}[t]
\centering
\includegraphics[width=0.98\textwidth]{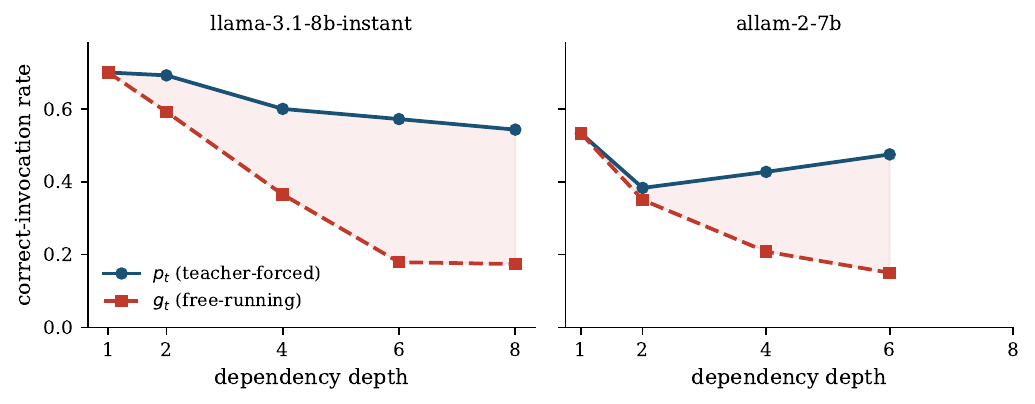}
\caption{$p_t$ (teacher-forced) vs.\ $g_t$ (free-running) by depth. The gap (shaded) is the propagation signature this study isolates; it is absent by construction at $d{=}1$ and widens with depth on both models.}
\label{fig:propagation-gap}
\end{figure}

Figure~\ref{fig:propagation-gap} and Table~\ref{tab:depth-results} show the core measurement. $p_t$ falls gently with depth ($0.700\to0.543$ on \texttt{llama-3.1-8b}, a 22\% relative decline from context growth) while $g_t$ collapses ($0.700\to0.174$, 76\%). $L_t$ rises monotonically with non-overlapping adjacent-depth intervals: the trend is resolved, not suggested.

\begin{table}[htbp]
\centering
\caption{Per-depth $p_t$, $g_t$, $L_t$, the two high-$n$ models.}
\label{tab:depth-results}
\footnotesize
\begin{tabular}{@{}lrrrl@{}}
\toprule
\multicolumn{5}{c}{\texttt{llama-3.1-8b-instant}} \\
\midrule
depth & $p_t$ & $g_t$ & $L_t$ [89\% CI] & $n_g$ \\
\midrule
1 & 0.700 & 0.700 & \textbf{0.000} $[0.000,0.000]$ & 60 \\
2 & 0.692 & 0.592 & 0.145 $[0.082,0.222]$ & 120 \\
4 & 0.600 & 0.365 & 0.391 $[0.311,0.474]$ & 260 \\
6 & 0.572 & 0.179 & \textbf{0.686} $[0.607,0.766]$ & 390 \\
8 & 0.543 & 0.174 & \textbf{0.680} $[0.574,0.772]$ & 184 \\
\midrule
\multicolumn{5}{c}{\texttt{allam-2-7b}} \\
\midrule
1 & 0.533 & 0.533 & \textbf{0.000} $[0.000,0.000]$ & 60 \\
2 & 0.383 & 0.350 & 0.087 $[0.022,0.163]$ & 120 \\
4 & 0.427 & 0.208 & 0.514 $[0.429,0.600]$ & 260 \\
6 & 0.475 & 0.150 & \textbf{0.684} $[0.624,0.745]$ & 240 \\
\bottomrule
\end{tabular}
\end{table}

\textbf{$L_1 = 0.000$ exactly on every model}: at depth 1 both arms build an identical prompt and share one cached response, so baseline purity holds by construction---the cheapest available check that the two arms are wired correctly. Step-index error counts on \texttt{llama-3.1-8b} at depth 6 show the same signature with no model at all: identical at step 0 (21/21), then the free arm rises to 40 while the teacher-forced arm stays near 20 at every later step. A flat teacher-forced profile rules out context growth as the explanation.

The model-free lag check $\Delta_1 = P(t{+}1\text{ ok}\mid t\text{ ok}) - P(t{+}1\text{ ok}\mid t\text{ wrong})$ is positive for every model with any errors ($+0.32$ to $+1.00$), confirming propagation with no parametric assumption; its second term is exactly zero for every model, so it reduces to $P(\text{ok}\mid\text{ok})$ here (explained in Section~\ref{sec:central-result}).

\subsection{Ceiling models, and where in a chain a model errs}
\label{sec:ceiling}

\begin{figure}[t]
\centering
\includegraphics[width=0.62\textwidth]{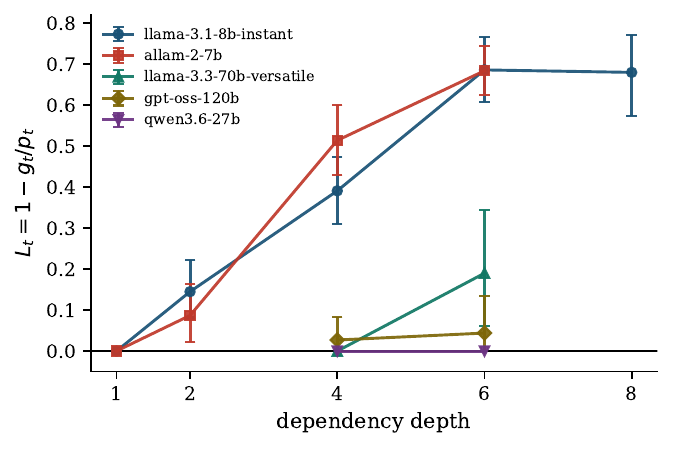}
\caption{$L_t$ vs.\ depth, all five usable models with 89\% CIs. Three models sit near zero at shallow depth; \texttt{llama-3.3-70b-versatile} is the only one to leave the ceiling with an interval excluding zero, at depth 6.}
\label{fig:lt-all}
\end{figure}

Three models sit at or near ceiling ($p_t=0.865$--$1.000$; Figure~\ref{fig:lt-all}), so $L_t\approx0$ there is a statement about the tasks, not robustness. On \texttt{llama-3.3-70b-versatile} at depth 4, every one of its 7 errors falls at step 2 or 3---the \emph{final} step, where nothing downstream exists to poison, so propagation was impossible there by construction. Depth 6 gave those errors somewhere to go: $L_6 = +0.190$ $[+0.062,+0.345]$, the first propagation loss \emph{established} in a large model here. \texttt{gpt-oss-120b} errs on a single fresh error and is not distinguishable from null at this $n$; \texttt{qwen3.6-27b} makes \textbf{zero errors across 298 free-arm steps}.

\textbf{An error on the final step cannot propagate}, so any fixed-depth measurement \emph{under-states} loss, worst where chains are shortest; $L_1{=}0$ is the limiting case of this bias. Terminal-error share is a \textbf{property of the model}, not depth, and no aggregate exposes it: on \texttt{llama-3.1-8b} it runs consistently above the uniform expectation (0.714 vs.\ 0.500 at $d{=}2$; 0.197 vs.\ 0.167 at $d{=}6$; $n=49$--$320$); \texttt{llama-3.3-70b-versatile}'s $d{=}4$ share of 6/7 is the vivid but weak case ($n{=}7$, CI $[0.488,0.992]$), and it is exactly why its $L_4$ read as immunity. \textbf{Practical consequence}: report terminal-error share alongside $L_t$, and read the depth \emph{trend}, since two models with identical per-call accuracy but different error position produce different---and non-comparable---propagation losses.

\subsection{The scoring regime, not the models, fixes both parametric quantities}
\label{sec:central-result}

This is the paper's central result, and it concerns a class of measurement, not the models we happened to measure.

\textbf{Severity is forced to its boundary.} $\pi \coloneqq 1 - P(\text{ok}\mid\text{poisoned})/P(\text{ok}\mid\text{clean})$ is a ratio of two directly observable rates, needing no fit. Measured: \textbf{0 of 869 poisoned-context steps correct}, so $\pi=1.000$. Support is uneven, stated as such: two models carry the claim (510, 335 poisoned steps, 95\% upper bounds 0.0059, 0.0089); two are consistent but uninformative (16, 8 steps, bounds 0.171, 0.312); \texttt{qwen3.6-27b} is \textbf{undefined}, not zero---it never left a clean context, so it has no severity to report, and pooling it as 0.000 would misleadingly read as ``poisoning does not hurt this model.''

\textbf{Recovery is structurally unobservable}: $P(\text{next ok}\mid\text{this wrong})=0.000$, \textbf{0 of 580} poisoned steps with a successor returned on-track. Both facts share one mechanism: a poisoned step holds a non-gold value, and the gold value at step $t$ is generated by tool constants \textbf{never exposed to the model}. After divergence, that value is information the model cannot derive; the only route back is coincidence at $\approx1/100{,}000$ per opportunity, giving an expected \textbf{0.0058} coincidental returns over 580 chances. Observing zero is thus almost uninformative: \textbf{a model that self-corrects 20\% of the time and one that never does produce identical observable data}, since neither can emit a value it has never seen. Severity is pinned at 1 and recovery unidentified for the same reason.

Substituting $\pi=1$, $r_{\mathrm{syn}}=r_{\mathrm{sem}}=0$ into Section~\ref{sec:propagation-model} leaves
\begin{equation}
g_t = c_t\,p_t, \qquad c_t = \textstyle\prod_{j<t} p_j, \qquad L_t = x_t = 1-c_t,
\end{equation}
\textbf{no free parameters at all}: an identity in measured per-step rates, not a non-identifiable fit. Run anyway, the three-parameter posterior returns \textbf{0.92 $[0.84,0.99]$} and \textbf{0.73 $[0.64,0.83]$} for a quantity exactly 1.000, fitting a smooth recurrence to pooled aggregates under a prior centred at 0.5. $\hat R\le1.0005$, ESS $\ge5{,}823$, zero divergences: \textbf{clean MCMC diagnostics certify the posterior was explored, and are silent on whether it was constrained.}

\textbf{Scope.} This depends only on scoring against a fixed gold trajectory whose values the model cannot reconstruct---i.e.\ execution-match and AST-match scoring generally. \textbf{Any study fitting a severity, self-correction, or recovery parameter under such scoring reports a parameter its scoring rule has already determined.}

\subsection{A remedy: conditional-on-state scoring}
\label{sec:remedy}

Credit a call when it \emph{correctly continues from the value the model actually holds}, not when it matches gold. Half already existed: routing-task selection was already scored both against gold and conditionally; we added the argument-side counterpart. Because the cache holds every raw completion, we applied this to data already collected---\textbf{0 API calls, 881 cache hits}---which is what makes it adoptable by any group holding cached completions.

\begin{table}[htbp]
\centering
\caption{Conditional-on-state severity vs.\ gold-agreement severity.}
\label{tab:conditional}
\footnotesize
\begin{tabular}{@{}lccc@{}}
\toprule
model & gold-agreement $\pi$ & conditional $\pi$ & 89\% CI \\
\midrule
\texttt{llama-3.1-8b-instant} & 1.000 (boundary) & \textbf{$+0.149$} & $[+0.021,+0.268]$ \\
\texttt{allam-2-7b} & 1.000 (boundary) & \textbf{$+0.316$} & $[+0.066,+0.503]$ \\
\bottomrule
\end{tabular}
\end{table}

Clean-context steps score identically under both rules (0.658 vs.\ 0.658), as they must, which is the implementation check. Both intervals exclude zero: poisoning \emph{does} measurably degrade rule application, so the strongest reading (``purely information loss'') is not supported. The supported reading is quantitative---gold-agreement attributes all loss to severity; conditional scoring attributes \textbf{0.15--0.32} to genuine degradation, the remainder to trajectory unreachability. We flag this as a hypothesis the data supports on two models with wide intervals, not a settled result: a correct argument here means transcribing an integer shown one turn earlier, a weak competence test, so this may not survive harder argument construction (a transformed-argument condition for this was built but \textbf{not run}; Section~\ref{sec:limitations}).

\textbf{Cost.} No single right answer per call; more implementation; looser cross-system comparability; a high correct-invocation rate stops implying end-task success. Exact-match buys comparability at the price of making severity/recovery unmeasurable; conditional scoring trades the opposite way. We report both, with gold-agreement as the primary headline.

\subsection{Rule-following is measurable directly, and aggregates hide it}
\label{sec:discrimination}

\begin{figure}[t]
\centering
\includegraphics[width=0.62\textwidth]{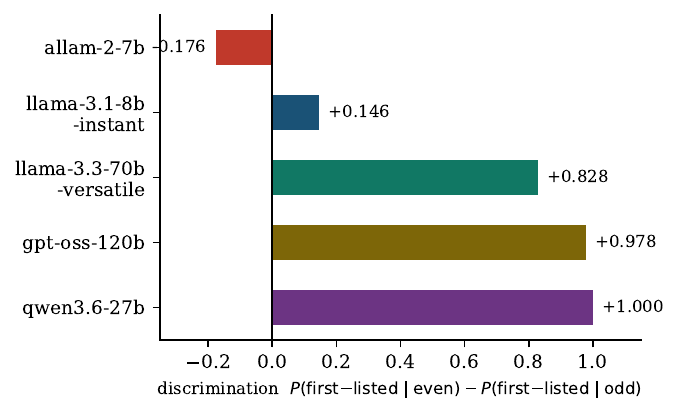}
\caption{Discrimination---$P(\text{first-listed}\mid\text{even})-P(\text{first-listed}\mid\text{odd})$---per model. Order matches the independently established reliability order of Section~\ref{sec:propagation-large}.}
\label{fig:discrimination}
\end{figure}

Define \textbf{discrimination} as $P(\text{first-listed}\mid\text{ref even})-P(\text{first-listed}\mid\text{ref odd})$: 0 means the rule is ignored, $+1$ perfect application, negative means \emph{anti}-correlated. Overall first-listed rates (0.478--0.497, four of five models) are indistinguishable and consistent with no position bias whatever; the \emph{same data}, conditioned on parity, resolves discrimination across the full range (Figure~\ref{fig:discrimination}), and the order matches reliability established independently in Section~\ref{sec:propagation-large}. \textbf{Discrimination is a per-model diagnostic of whether a model performs the task at all, invisible to any metric aggregated over the conditioning variable.}

Three qualitatively different relationships, not one spectrum: \texttt{qwen3.6-27b} \emph{derives} the rule ($+1.000$, zero errors in 298 steps, explicit reasoning trace); \texttt{llama-3.1-8b} \emph{weakly follows} it ($+0.146$, $z{=}4.7$); \texttt{allam-2-7b} is \emph{anti-correlated} ($-0.176$, $z{=}-5.0$)---it picks the first-listed tool \emph{more} on odd refs (0.774) than even (0.597), where its accuracy is worst (0.223 vs.\ 0.595). \texttt{allam-2-7b}'s propagation loss is therefore not ``a weak model that errs more'' but a model \emph{not performing the routing task}.

\subsection{Two negative results}

\textbf{Untestable, not underpowered:} a hypothesised error-composition shift along the chain cannot be tested here because \textbf{389 of 389} clean-context errors are selection errors---on a copy-argument task there are only two ways to fail cleanly, mis-apply the parity rule (hard) or mis-transcribe a verbatim number (models never do). \textbf{Unresolved by design}: the suite offers two 2-point scale contrasts rather than 3--4, not comparable in kind (\texttt{llama} 8B$\to$70B dense, $8.75\times$; \texttt{gpt-oss} 20B$\to$120B, $6\times$ total but $\approx1.4\times$ active params). We report both axes and draw no scale conclusion, since the larger models sit at ceiling on these tasks.

\section{Pipeline Validation, and What Simulation Cannot Establish}
\label{sec:validation}

Before spending API budget we validated the full pipeline against six simulated policies with known ground truth, eliminating seven harness artifacts (retry-budget asymmetry, selection errors mis-attributed under poisoning, a contaminated syntactic-share estimator, among others). Two further artifacts were not coding errors, which is what makes them instructive: comparing carried value against \emph{expected argument} rather than \emph{previous gold output} is correct on a copy-argument task and silently wrong under any transform; and the simulated policies recovered by reading ground truth directly, out of band, through a channel no real model has. Correct code, correct configuration, invalid measurement---and the second artifact is why the recovery channel was never estimable from observable data in any version of this pipeline.

Three principles follow, nested by scope: \textbf{(1) simulated validation} certifies only the channels it exercises and silently passes the ones it does not (our mock backend read state from stashed context rather than the message history, so an agent loop that never threaded observations into the conversation passed the entire simulated suite); \textbf{(2) a parser or scorer can silently flip the \emph{direction} of a result}, not just its precision---replaying cached completions through our pre-fix extractor, \textbf{319/394 (81.0\%)} of \texttt{qwen3.6-27b}'s responses would have failed to parse; \textbf{(3) every constant estimated from one model or assumed for convenience should be verified per-model}---a shared output-token constant hid \texttt{qwen}'s true cost (6.6$\times$ higher); an assumption that parity was incidental hid \texttt{allam}'s anti-correlation; per-depth pooling hid where each model's errors cluster.

This establishes that the estimation and aggregation pipeline recovers configured parameters. It does \textbf{not} establish the prompt-construction and observation-passing path, and no out-of-band simulation can.

\section{Limitations and Scope}
\label{sec:limitations}

\textbf{This study is a controlled, synthetic-task measurement, not a validation on production tool-calling benchmarks.} Integer-argument tasks give exact ground truth and let us isolate propagation cleanly, at the cost of ecological validity: whether these findings---particularly the conditional-scoring decomposition of Section~\ref{sec:remedy}, where a correct argument means transcribing a value shown one turn earlier---hold under natural-language or multi-field arguments is open. Extending this protocol to BFCL- or tau-bench-style tasks, where gold values are not perfectly opaque tool outputs, is the direct next step and would additionally test whether the identifiability result of Section~\ref{sec:central-result} is a property of exact-match scoring in general or specific to opaque-constant synthetic tasks.

\textbf{Four designed conditions were not executed within this study's compute budget}, each a scope boundary rather than a missing result: a linear-task null control (Section~\ref{sec:propagation-large}'s evidence for task-structure specificity therefore rests on internal signatures---flat teacher-forced profile, exact $L_1{=}0$, step-index divergence---rather than a direct null arm); a transformed-argument condition (leaves the error-composition hypothesis and the conditional-scoring generality question of Section~\ref{sec:remedy} open); a presentation-order control (the parity confound of Section~\ref{sec:discrimination} is addressed by the discrimination statistic alone, not removed by design); and a calling-mode ablation (native mode was verified feasible on 5 of 6 models but not run at scale, so we make no claim about how much measured unreliability is a calling-mode artifact). We name these individually because a paper reporting effect sizes should be explicit about which arms produced them.

\textbf{Other scope notes.} Soft argument matching coincides with strict matching here (single-integer arguments); the opaque-feedback condition is implemented but unrun, so feedback-format hypotheses are not addressed; \texttt{distractor\_level} does not vary on routing tasks, so selection rates are not comparable across the routing and linear arms; interval widths differ substantially across models by design (nested unequal $n$), so no scale or family contrast should be read from point estimates alone.

\section{Conclusion}

We measured tool-use reliability at the invocation level, separating tool selection from argument correctness and context-length decay from error propagation, and found propagation dominant on small models: by depth 6, roughly 70\% of a model's own clean-context capability is lost to its own earlier mistakes.

The unexpected result concerns measurement. Under exact-match scoring against a fixed gold trajectory, a propagation model's severity and recovery parameters are not weakly identified---they are \textbf{determined in advance by the scoring rule}, because after divergence the gold value is information the model has never received and cannot derive. The recurrence collapses to an identity in measured per-step rates; a fit run anyway returns confident, wrong numbers while every convergence diagnostic reports health. The remedy is a scoring change, not a modelling one: credit a call that correctly continues from the value the model actually holds, computable retrospectively from cached completions at zero marginal cost, which moves severity from a boundary artifact to interior estimates excluding zero.

For agent benchmarks scoring against reference trajectories generally: \textbf{if the scoring rule makes the reference unreachable after a divergence, any severity or recovery parameter fit under it is a property of the scorer, not the model.}

{\footnotesize
\bibliographystyle{plain}

}

\end{document}

%% file: methodology_tikz.tex
\begin{tikzpicture}[
    box/.style={draw, rounded corners=2pt, minimum height=0.65cm, align=center, font=\footnotesize, inner sep=4pt},
    state/.style={draw, circle, minimum size=0.9cm, font=\scriptsize, align=center},
    arrow/.style={-{Latex[length=2mm]}, thick},
    dashedarrow/.style={-{Latex[length=2mm]}, dashed},
    scale=0.92, every node/.style={transform shape}
]

\node[box, fill=blue!8, minimum width=3.1cm] (task) at (0,3.4) {task at depth $t$\\(schema + history)};
\node[box, fill=green!10, minimum width=3.4cm] (tf) at (-2.6,2.0) {teacher-forced arm\\clean gold history};
\node[box, fill=red!10, minimum width=3.4cm] (free) at (2.6,2.0) {free-running arm\\model's own history};
\node[box, minimum width=2.6cm] (pt) at (-2.6,0.7) {measure $p_t$};
\node[box, minimum width=2.6cm] (gt) at (2.6,0.7) {measure $g_t$};
\node[box, fill=yellow!15, minimum width=3.4cm] (Lt) at (0,-0.6) {$L_t = 1 - g_t/p_t$\\net propagation loss};

\draw[arrow] (task) -- (tf);
\draw[arrow] (task) -- (free);
\draw[arrow] (tf) -- (pt);
\draw[arrow] (free) -- (gt);
\draw[arrow] (pt) -- (Lt);
\draw[arrow] (gt) -- (Lt);

\begin{scope}[yshift=-0.45cm]
\node[state, fill=green!15] (C) at (7.3,2.6) {clean\\$c_t$};
\node[state, fill=orange!15] (S) at (10.3,3.6) {poisoned\\syntax\\$s_t$};
\node[state, fill=red!15] (M) at (10.3,1.6) {poisoned\\semantic\\$m_t$};

\draw[arrow] (C) to[bend left=12] node[above,font=\tiny,sloped]{$(1{-}p_t)f_{\mathrm{syn}}$} (S);
\draw[arrow] (C) to[bend right=12] node[below,font=\tiny,sloped]{$(1{-}p_t)(1{-}f_{\mathrm{syn}})$} (M);
\draw[arrow] (S) to[bend left=12] node[above,font=\tiny,sloped]{$r_{\mathrm{syn}}$} (C);
\draw[arrow] (M) to[bend right=12] node[below,font=\tiny,sloped]{$r_{\mathrm{sem}}$} (C);
\draw[arrow] (C) to[loop above] node[above,font=\tiny]{$p_t$} (C);

\node[font=\scriptsize, align=center, text width=3.4cm] at (8.8,-0.2) {\textbf{Under exact-match scoring}\\(Sec.~5.3): $\pi \to 1$, $r_{\mathrm{syn}}, r_{\mathrm{sem}} \to 0$\\forced, not fitted};
\end{scope}

\node[font=\bfseries\small] at (0,4.1) {A.\ Two-arm scoring protocol};
\node[font=\bfseries\small] at (8.8,4.35) {B.\ Propagation model};

\end{tikzpicture}